\documentclass[letterpaper, 10 pt, conference]{ieeeconf}  

\IEEEoverridecommandlockouts                              

\usepackage{graphicx} 
\usepackage{amsmath} 
\usepackage{amssymb}  
\usepackage{soul}
\usepackage{array}
\usepackage[dvipsnames]{xcolor}
\usepackage{wrapfig}
\usepackage{caption}
\usepackage{multicol}
\usepackage{subcaption}
\let\labelindent\relax  
\usepackage{enumitem}
\usepackage{algorithm}
\usepackage[noend]{algpseudocode}
\usepackage{bbding}
\usepackage{pifont}
\usepackage{wasysym}
\usepackage{capt-of,etoolbox}
\usepackage[bookmarks=false]{hyperref}

\makeatletter
\def\BState{\State\hskip-\ALG@thistlm}
\patchcmd\@maketitle\null{{\myfigure{}\par}}{}{}
\makeatother

\newenvironment{flushitemize}{
  \begin{itemize}[leftmargin=*, labelsep=0.5em]
    \setlength{\itemindent}{0pt}
    \setlength{\labelwidth}{0pt}
    \setlength{\labelsep}{0.5em}
}{\end{itemize}}

\title{\LARGE \bf
Many-to-Many Multi-Agent Pickup and Delivery
}

\author{Ethan Schneider$^{1}$, Jingkai Chen$^{2}$, Tianyi Gu$^{2}$, Kunlei Lian$^{2}$, Seth Hutchinson$^{3}$, Sonia Chernova$^{1}$
\thanks{This work is sponsored by Symbotic.}
\thanks{$^{1}$Institute of Robotics and Intelligent Machines, Georgia Institute of Technology, Atlanta, GA 30332, USA \texttt{eschneider32@gatech.edu}}%
\thanks{$^{2}$Symbotic Inc., Wilmington, MA 01887, USA}%
\thanks{$^{3}$Northeastern University, Boston, MA 02115, USA}%
}

\begin{document}

\maketitle
\thispagestyle{empty}
\pagestyle{empty}

\begin{figure*}[h!]
    \centering
    \includegraphics[width=0.93\linewidth]{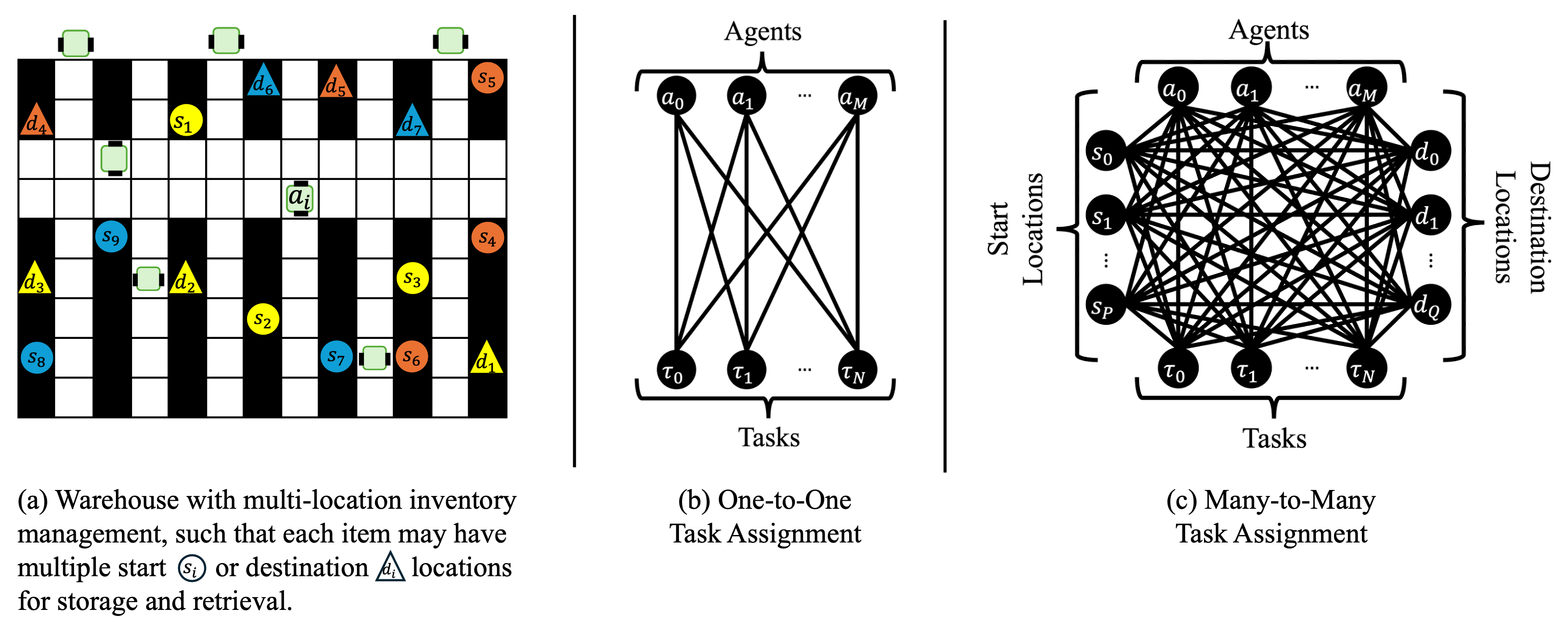} 
    \caption{(a) Multi-location inventory management is common practice in warehouse settings, where multiple source and destination locations are available for storing and retrieving goods. In multi-agent pickup and delivery contexts, the choice of source and destination locations can significantly impact system performance (e.g., for agent $a_i$ (center above) the sequence $s_3 d_1 s_4 d_5$ is far more efficient than $s_1 d_3 s_4 d_4$). (b) Current state-of-the-art methods focus on one-to-one allocation between agents and tasks, ignoring multi-location inventory options. (c) We introduce Many-to-Many Multi-Agent Pickup and Delivery, which accounts for agents, tasks, and multiple source and destination locations per item. Our computationally-efficient approach yields significant improvements in task throughput in long-horizon simulations across 3 warehouse layouts.}
    \vspace{-1em}
    \label{fig:o2o-m2m}
\end{figure*}

\begin{abstract}
Multi-robot systems in automated warehouses must manage continuous streams of pickup-and-delivery tasks while ensuring efficiency and safety. Prior work on Multi-Agent Pickup-and-Delivery (MAPD) has largely focused on the one-to-one variant, where each task has a fixed pickup and delivery location. In contrast, real warehouses often present many-to-many MAPD scenarios, where items, tracked by stock keeping unit (SKU) identifiers, can be retrieved from or stored at multiple locations, resulting in an NP-hard four-dimensional assignment problem. To solve the many-to-many MAPD problem, we contribute our algorithm: Many-to-Many Multi-Agent Pickup and Delivery (M2M). We experiment with two variants of our algorithm: one that minimizes estimated task durations (M2M), and one which incorporates SKU distribution into the objective function (M2M-wSKU). Simulation results over 8-hour warehouse operations show that our method consistently matches or outperforms prior state of the art, with M2M completing up to 22,000 more tasks on average across different environments and warehouse inventory densities.

\end{abstract}

\section{Introduction} \label{sec:introduction}

In many multi-robot systems, agents must continuously manage incoming tasks while planning collision-free routes. A common formulation for this problem is that of Multi-Agent Pickup-and-Delivery (MAPD) \cite{ma2017lifelong}, where each task specifies a pickup location and a corresponding delivery location. The most widely studied case, in which each task has exactly one pickup and one delivery, is referred to as the \textit{one-to-one pickup-and-delivery problem} \cite{berbeglia2010dynamic}. To date, nearly all MAPD approaches have focused on this one-to-one setting, typically casting task allocation as a two-dimensional assignment problem. These works then employ classical optimization techniques such as the Hungarian Algorithm \cite{kuhn1955hungarian} or reformulations into related combinatorial problems such as the Traveling Salesman Problem \cite{liu2019task}.

In real-world settings, such as automated warehouses, inventory often contains many identical items, tracked by a stock keeping unit (SKU) identifier (Figure \ref{fig:o2o-m2m}). In these settings, fulfilling a request typically means retrieving any available instance of the item (e.g., any 28lb bag of Famous Brand Dog Food among many stored in the warehouse), and newly arriving items may similarly be stored in any of several valid locations. This variant of the problem is known as \textit{many-to-many pickup-and-delivery} \cite{berbeglia2010dynamic}. Unlike the one-to-one setting where each task specifies a fixed pickup and delivery location, agents must now jointly decide which task to execute and which pickup and delivery sites to use. This increases the complexity of task allocation, which can be formulated as a four-dimensional assignment problem, a known NP-hard problem \cite{gabrovvsek2020multiple}.

Our goal is to achieve consistently high performance over long time horizons in automated warehouse environments by addressing the many-to-many multi-agent pickup-and-delivery problem. In this setting, task allocation requires jointly assigning agents, tasks, pickup locations, and delivery destinations while accounting for factors such as task duration and the spatial distribution of SKUs. In contrast, prior work on MAPD has largely focused on the one-to-one variant, where pickup and delivery locations are fixed or selected randomly \cite{lnspbs, ma2017lifelong, chen2021integrated}, without modeling the dynamics of an evolving inventory. Our work shows that ignoring the many-to-many assignment problem and item distribution can lead to inefficient allocations and higher performance variance, underscoring the need for approaches that explicitly reason about these factors.

In this work, we introduce our sequential many-to-many MAPD algorithm, \textit{Many-to-Many Multi-Agent Pickup and Delivery (M2M)}. M2M first computes an initial suboptimal many-to-many allocation over agents, tasks, start locations, and destinations, and then iteratively improves the allocation using the Large Neighborhood Search (LNS) metaheuristic \cite{lnssurvey} for a pre-specified time. We then input the final allocation to the multi-agent path finding solver Priority Based Search (PBS) \cite{ma2019searching}. We evaluate two variants of our algorithm which differ in their cost function: M2M minimizes the sum of estimated task durations, encouraging agents to complete tasks sooner, whereas our other variant, \textit{M2M with SKU Distribution (M2M-wSKU)}, incorporates the distribution of SKUs in the environment into the objective function, promoting improved item placement for both current and future system performance. We evaluate our methods against the MAPD baseline LNS-PBS \cite{lnspbs} in an online setting using 8-hour warehouse simulations across three layouts and three inventory densities to assess long-term performance. Our results show that M2M consistently matches or outperforms LNS-PBS and M2M-wSKU in task throughput on all layouts and inventory densities. Over an 8-hour simulation window, M2M completes up to 22,000 more tasks on average than LNS-PBS, representing an increase of up to 38.8\% depending on warehouse layout and inventory density.

\section{Related Works} \label{sec:related works}
The MAPD problem can be broken into the two subproblems, Multi-Robot Task Allocation (MRTA) and Multi-Agent Path Finding (MAPF). We frame our work in the context of these three subareas below.

\subsection{Multi-Robot Task Allocation (MRTA)}

The MRTA problem consists of deciding which tasks are assigned to which agents such that the system's goals are achieved. Gerkey et al. \cite{gerkeysurvey} provide a MRTA taxonomy for whether a robot is capable of working a single task (ST) or multiple tasks (MT), whether a task requires exactly one robot (SR) or allows multiple robots to work simultaneously (MR), and whether only an instantaneous allocation (IA) or a sequence of allocations into the future is allowed (TA). In this work, we focus on the ST-SR-TA variant, a known NP-Hard problem \cite{gerkeysurvey}. Korsah et al. \cite{korsahsurvey} later provide an extensive survey on the MRTA problem, which focuses on the interrelatedness and complexity of tasks in the system. In this work, we place ourselves in the No Dependencies (ND) ST-SR-TA problem. 

Prior work exists to compute the task allocation for the one-to-one pickup-and-delivery problem. The Hungarian Algorithm \cite{kuhn1955hungarian} solves the corresponding assignment by finding a maximum weight bipartite matching in polynomial time, while another method is to formulate and solve a corresponding Traveling Salesman Problem (TSP) \cite{liu2019task}. Many-to-many pickup-and-delivery is modeled as a four-dimensional assignment problem. Prior work for solving these higher dimension problems has introduced offline methods by modifying the Hungarian Algorithm for dimensions greater than two \cite{gabrovvsek2020multiple}, utilizing greedy-like algorithms \cite{gutin2008greedy}, or formulating a mixed-integer linear program (MILP) \cite{walteros2014integer}. These methods are part of the MAPD solution, which we discuss in Subsection \ref{subsec: MAPD}. 

\subsection{Multi-Agent Path Finding (MAPF)}

The multi-agent path finding (MAPF) problem is to compute collision-free paths for multiple agents from their start position to one or more destinations per agent. Prior work introduces the optimal and complete algorithm Conflict Based Search (CBS) \cite{sharon2015conflict}, which independently plans single-agent path planning, then resolves agent conflicts via a high-level search in a constraint tree. Enhanced Conflict Based Search (ECBS) \cite{barer2014suboptimal} addresses the scalability issue of CBS while employing a suboptimal bound. Cooperative A* \cite{silver2005cooperative} employs a prioritized planning approach, in which agents compute path plans in order of priority. Ma et al. \cite{ma2019searching} investigate the theoretical limitations of prioritized planning and present Priority Based Search (PBS), which explores the space of all priority orderings of agents using a systematic depth-first search. Grenouilleau et al. propose MLA*, which \cite{grenouilleau2019multi} extended prioritized planning from a single goal location to pairs of goal locations, namely a start and destination in pickup-and-delivery problems. Li et al. \cite{li2021lifelong} generalize MLA* to handle longer sequences of goal locations by utilizing a planning window approach with their algorithm Rolling Horizon Collision Resolution (RHCR). Lastly, recent work by Jiang et al. \cite{jiang2025deploying} utilizes imitation learning techniques to learn and scale prior search based MAPF solvers. 

\begin{table}
\begin{tabular}{c|c|c|c}
\hline
Method & online & \begin{tabular}[c]{@{}c@{}}assign seq.\\ of tasks\end{tabular} & \begin{tabular}[c]{@{}c@{}}many-to-many\end{tabular} \\ \hline
CENTRAL \cite{ma2017lifelong} & \ding{51} & \ding{55} & \ding{55}  \\ \hline
TA-Hybrid \cite{liu2019task} & \ding{55} & \ding{51}  & \ding{55} \\ \hline
HBH+MLA* \cite{grenouilleau2019multi} & \ding{51} & \ding{55} & \ding{55} \\ \hline
RMCA \cite{chen2021integrated} & \ding{51} & \ding{51} & \ding{55} \\ \hline
LNS-PBS \cite{lnspbs} & \ding{51} & \ding{51} & \ding{55} \\ \hline
M2M / M2M-wSKU (ours) & \ding{51} & \ding{51} & \ding{51} \\ \hline
\end{tabular}
\caption{Summary of Prior MAPD methods. ``Online" indicates that not all tasks are known a priori and tasks arrive over time. ``Assign seq. of tasks" means agents are assigned a sequence of tasks rather than a single task. ``Many-to-many" distinguishes algorithms that solve tasks with multiple available start and destination locations per task from those assuming a single start and destination (one-to-one).}
\label{tab:prior-works}
\vspace{-1em}
\end{table}

\subsection{Multi-Agent Pickup and Delivery (MAPD)} \label{subsec: MAPD}

Table \ref{tab:prior-works} summarizes recent works addressing the MAPD problem. Prior work by Ma et al. proposed CENTRAL \cite{ma2017lifelong}, which uses the Hungarian Algorithm to solve the MRTA problem and CBS for planning collision-free paths. TA-Hybrid \cite{liu2019task} solves the offline MAPD problem by formulating a Traveling Salesman Problem, as discussed earlier, then uses a mixed MAPF approach of ICBS \cite{barer2014suboptimal} and a min-cost max-flow algorithm depending on the number of agent groups. HBH-MLA* \cite{grenouilleau2019multi} utilizes a h-value-based heuristic for task allocation while prioritized planning and MLA* solve the MAPF problem. Work by Chen et al. proposed the simultaneous MAPD algorithm regret-based marginal-cost assignment (RMCA) \cite{chen2021integrated}, which utilizes LNS informed by actual-path costs for MRTA and prioritized planning with A* for MAPF. Building on these earlier insights, recent work by Xu et al. \cite{lnspbs} proposed LNS-PBS to solve the MRTA problem by using the Hungarian Algorithm for an initial solution, which is then improved upon via Large Neighborhood Search \cite{lnssurvey}, followed by utilizing PBS to address the MAPF problem. The authors empirically show that their algorithm outperforms all other MAPD approaches listed in Table \ref{tab:prior-works}; we therefore utilize LNS-PBS as our baseline method as discussed in Section \ref{sec:experimental setup}. 

\section{Problem Formulation} \label{sec:problem formulation}
\begin{figure*}[t]
    \centering
    \includegraphics[width=1.0\linewidth]{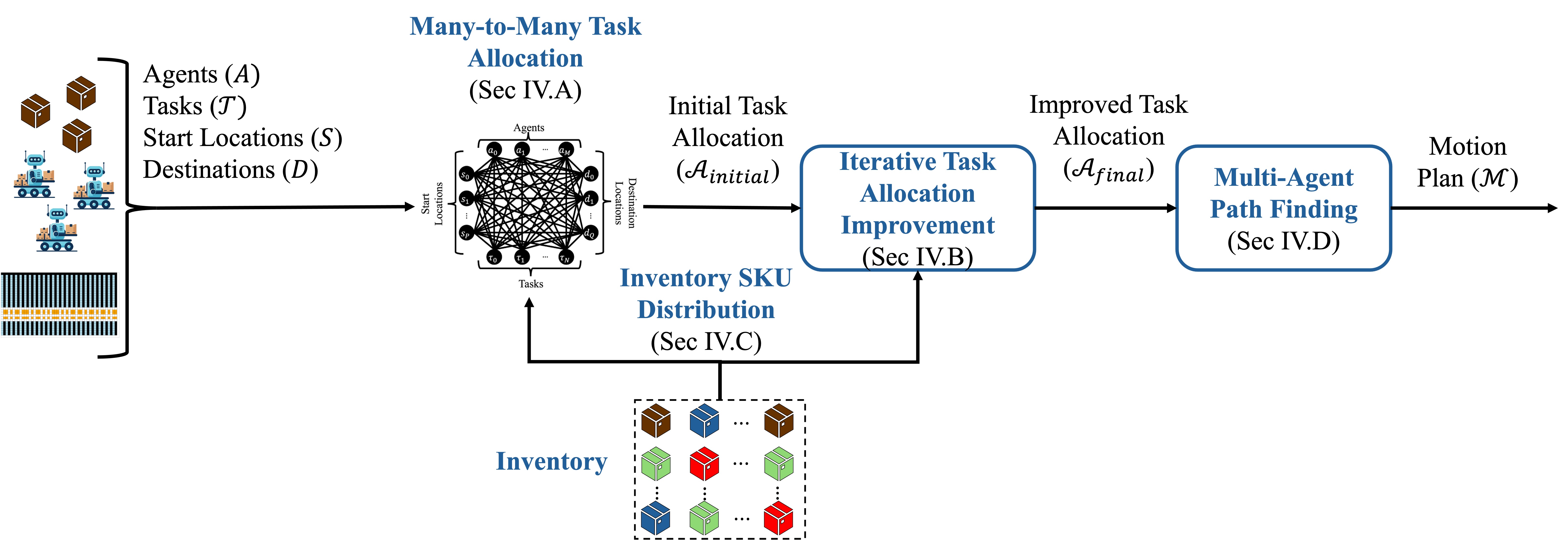}
    \caption{Overview of M2M method and M2M-wSKU variant.}
    \vspace{-1em}
    \label{fig:system-diagram}
\end{figure*}



We define the Many-to-Many Multi-Agent Pickup and Delivery (M2M-MAPD) problem as follows: given a set of agents $A$ and a set of tasks $\mathcal{T}$, where each task specifies a source location and a delivery location, our objective is to compute task allocation $\mathcal{A}$ and a collision-free motion plan $\mathcal{M}$ such that we maximize the overall performance of the system measured as task throughput (tasks/min). In the many-to-many setting, agents are allocated to tasks, where each agent chooses the task's start and destination locations from multiple task-specific options. This many-to-many variant can be modeled as a 4-dimensional assignment problem involving agents, tasks, start locations, and destinations, aiming to minimize the sum of costs of the task assignments. In contrast, one-to-one settings assign agents to tasks using predefined start and destination locations, resulting in a 2-dimensional assignment problem that minimizes the sum of assignment costs between agents and tasks.


Formally, we define a set of $M$ agents $A=\{a_0, a_1, ..., a_M\}$, each capable of moving through a shared environment represented as an undirected graph $G=(V, E)$ in which vertices $v_i \in V$ represent locations and edges $e_{ij} \in E$ represent traversable paths between vertices. The location of agent $a_i$ at timestep $t$ is denoted by $l_i(t) \in V$, and at each timestep, agents either move to an adjacent location or wait at their current location. Motion of the agent must satisfy the feasibility constraints encoded by the graph G and exclude all positions that result in collision, where a collision between agents $a_i$ and $a_j$ is defined as being co-located on the same vertex ($l_i(t)=l_j(t)$) or the same edge ($l_i(t)=l_j(t+1)\ \& \ l_i(t+1)=l_j(t)$).

We define $\mathcal{T}$ as the full set of robot tasks, composed of the union of already allocated and unallocated tasks, $\mathcal{T} = \{\mathcal{T}_{alloc} \cup \mathcal{T}_{free} \}$. In a warehouse context, we refer to tasks that bring items in and out of the warehouse as inbound ($\mathcal{T}_{in}$) and outbound ($\mathcal{T}_{out}$) tasks, respectively. Each task $\tau_n \in \mathcal{T}$ is specified by the tuple $\tau_n=(S_n, D_n)$, where $S_n \subseteq V$ represents the set of possible start locations for task $\tau_n$ and $D_n \subseteq V$ represents the set of possible destinations for task $\tau_n$. Task $\tau_n$ being allocated to agent $a_m$ is denoted by $\tau_n^m = (s_p, d_q)$, with select start location $s_p \in S_n$ and destination $d_q \in D_n$. The set of start locations and set of destinations for all tasks are defined as $S = \cup^N_{n=0}S_n$ and $D=\cup^N_{n=0}D_n$, respectively, such that $|S|=P$ and $|D|=Q$. 

A solution to a M2M-MAPD problem takes the form of task allocation $\mathcal{A}$ and a collision-free motion plan $\mathcal{M}$ that meets task ordering constraints such that $s_p$ is visited prior to $d_q$. An allocation $\mathcal{A}$ may assign an agent up to $k$ tasks in sequence, such that $\mathcal{T}^m=[\tau_0^m, \tau_1^m, ..., \tau_k^m]$. An agent with an empty task sequence is referred to as a free agent, otherwise it is referred to as an active agent. Our objective is to maximize the overall performance of the system measured as task throughput (tasks/min) while maintaining computational cost below some domain-specific threshold (e.g., 1 second).

\section{Methods} \label{sec:methods}

Fig. \ref{fig:system-diagram} presents an overview of our method: \textit{Many-to-Many Multi-Agent Pickup and Delivery (M2M)} which computes a many-to-many task allocation and corresponding multi-agent motion plan. The input to M2M is the set of agents ($\mathcal{A}$), and the set of tasks ($\mathcal{T}$) with their corresponding start ($S$) and destination ($D$) locations. As presented in Section \ref{sec:m2m-ta}, we contribute a novel approach for solving the 4-dimensional task assignment problem that results in an initial many-to-many task allocation $\mathcal{A}_{initial}$. Inspired by prior work \cite{lnspbs}, we then utilize the anytime algorithm Large Neighborhood Search (LNS) to iteratively improve $\mathcal{A}_{initial}$, resulting in the best-found solution $\mathcal{A}_{final}$ (Section \ref{sec:lns}), which we then use to compute a motion plan as described in Section \ref{sec:mapf}.  In Section \ref{sec:sku} we present the \textit{M2M with SKU Distribution (M2M-wSKU)} variant of our approach that further improves performance by reasoning about spacial item distribution when selecting source and destination locations.



\subsection{Many-to-Many Task Allocation}
\label{sec:m2m-ta}

Our algorithm, M2M, is initiated when the system has unassigned tasks ($\mathcal{T}_{free}\neq \emptyset$).  M2M first computes an initial task allocation $\mathcal{A}_{initial}$, by iteratively assigning the minimum cost agent, task, start location, destination pairing until all tasks, start, or destinations have been assigned. To encode allocation costs, we construct a cost tensor $C\in\mathbb{R}^{M\times N \times P \times Q}$, in which an element $c_{(m,n,p,q)}\in C$ is the estimated cost of traveling from the last destination in $\mathcal{T}^m$ to start location $s_p$ to destination $d_q$.
In addition to the estimated duration cost, $C$ must encode whether $s_p \in S_n$ and $d_q \in D_n$, and whether $s_p \in \mathcal{T}_{alloc}$ and $d_q \in \mathcal{T}_{alloc}$. We define the cost function as: 
\[
c_{m, n, p, q}=\begin{cases}
    K & \text{if } s_p \notin S_n \\
    K & \text{if } d_q \notin D_n \\
    K & \text{if } s_p \in \mathcal{T}_{alloc} \\
    K & \text{if } d_q \in \mathcal{T}_{alloc} \\
    K & \text{if } |\mathcal{T}^m|>=\beta \\
    cost(a_m, s_p) + cost(s_p, d_q) & \text{otherwise}
\end{cases}
\]
where $K$ is a very large cost (e.g., $inf$) that corresponds to an invalid task allocation. $K$ is assigned when the start location is not in the set of possible start locations for task $\tau_n$ ($s_p \notin S_n$), the destination is not in the set of possible destinations for task $\tau_n$ ($d_q \notin D_n$), or if the start location or destination is already assigned to another task, $s_p \in V_{alloc}$ and $d_q \in V_{alloc}$, respectively. In this work, we limit the length of each agent's task sequence $\mathcal{T}^m$ to $\beta$, such that when $|\mathcal{T}^m|>=\beta$, the cost is $K$. Otherwise, the cost of $c_{m,n,p,q}$ is the estimated duration of agent $a_m$ traveling from the final destination of $\mathcal{T}^m$ to $s_p$, denoted by $cost(a_m, s_p)$, and the estimated duration to traverse from $s_p$ to $d_q$ is denoted by $cost(s_p, d_q)$.

For the smallest sized problem we investigate, i.e. $M=40, N=120, P=100, Q=200$, the number of elements in $C\approx 10^8$, which makes constructing, modifying, and searching $C$ computationally expensive with poor scalability to larger environments, more agents, and more tasks. 
To address this issue, we can decompose $C$ by relating parts of the cost function with edge weights between pairs of sets. We then decompose $C$ into four cost matrices,
$C_{AS}\in  \mathbb{R}^{M\times P}$, $C_{SD}\in \mathbb{R}^{P\times Q}$, $C_{\mathcal{T}S}\in \{0, 1\}^{N\times P}$, $C_{\mathcal{T}D}\in \{0, 1\}^{N\times Q}$. In which, $C_{AS}$ and $C_{SD}$ correspond to the estimated duration of agent $a_m$ traveling from its final destination in $\mathcal{T}^m$ to $s_p$, denoted as $cost(a_m, s_p)$, and the estimated duration traveling from $s_p$ to $d_q$, denoted $cost(s_p, d_q)$. Matrices $C_{\mathcal{T}S}$ and $C_{\mathcal{T}D}$ encode what start locations and destinations are valid for what tasks, in which 0 or 1 represents the location being invalid or valid respectively. 

The four cost matrices are used to compute an initial suboptimal assignment of agents, tasks, start locations, and destinations. Prior to task allocation, we add every task in each agent's task sequence to $\mathcal{T}_{free}$ except for the first task in the sequence, i.e. the task currently being worked by the agent. To avoid reconstructing $C$ for each allocation, we iterate over each task $\tau_n \in \mathcal{T}$ while keeping track of the best found allocation and cost. For each $\tau_n$, the valid start locations $C_{\mathcal{T}S}[n, :]$ and valid destinations $C_{\mathcal{T}D}[n, :]$ are combined with $C_{AS}$ and $C_{SD}$ to construct $C_{ASD} \in \mathbb{R}^{M\times P \times Q}$. The minimum element $c_{m,p,q}\in C_{ASD}$ is found, and if smaller than the best, is kept. After iterating through all tasks, the best element $c_{m,n,p,q}$ is allocated. Once allocated, cost matrices $C_{AS}[m,:]$ is updated to reflect the new final destination of agent $a_m$, while $C_{\mathcal{T}S}[n, :]$ and $C_{\mathcal{T}D}[n, :]$ are set to 0, so that $\tau_n$ will not be allocated again. Additionally, $C_{\mathcal{T}S}[:, p]$ and $C_{\mathcal{T}D}[:, q]$ are set to 0, so the start and destinations are not allocated to different tasks. This process is repeated until either $\mathcal{T}_{free}=\emptyset$ or every element in $C_{\mathcal{T}S}$ or $C_{\mathcal{T}D}$ is equal to 0, at which point we obtain $\mathcal{A}_{initial}$.

\subsection{Large Neighborhood Search (LNS)}
\label{sec:lns}

Given $\mathcal{A}_{inital}$, we use LNS to iteratively attempt to improve the current solution $\mathcal{A}_{curr}$ until a specified time is reached, when LNS returns the best found solution, $\mathcal{A}_{final}$. During the improvement process, LNS will first destroy a portion of the current solution via a destroy operator, then reassign the removed tasks via a repair operator. The new solution $\mathcal{A}_{new}$ is accepted as $\mathcal{A}_{curr}$ with a metropolis acceptance criterion \cite{kirkpatrick1983optimization, ropke2006adaptive}. The details of our LNS implementation are provided in the following subsections. 

\subsubsection{Shaw Removal}

We apply the Shaw Removal operator \cite{shaw1997new, shaw1998using} to remove $N_{remove}$ interrelated tasks from the current solution. First, an allocated task is randomly chosen to be removed, $\tau^*$. We then compute the interrelatedness between $\tau^*$ and every other task, except for the first task in each $\mathcal{T}^m$, since we do not want to interrupt an agent currently working on a task. Interrelatedness is defined as: 

\[r(\tau_i, \tau_j)= \omega_1(dist(d_i, d_j) + dist(s_i, s_j)) +\]
\[\omega_2(|t(s_i)-t(s_j)|+|t(d_i)-t(d_j)|)\]

Here, $\omega_1$ is the weight for spatial distance between two tasks' start locations, $dist(s_i, s_j)$, and destinations $dist(d_i, d_j)$, where $dist$ is the L1 norm. The second weight, $\omega_2$, is for the temporal relatedness of two tasks, in which $t(s_i)$ and $t(s_j)$ are the estimated times to reach the start locations and destinations for $\tau_i$ respectively. 

Once the relatedness scores between $\tau^*$ and all other tasks are computed, the top $N_{remove}$-1 most related tasks are added to $\mathcal{T}_{free}$ along with $\tau^*$. Additionally, for each $\mathcal{T}^m$, we remove any subsequent tasks to any removed via the Shaw Removal operator. The reasoning is that a task was appended in part to the agent's prior destinations in its task sequence, so if a task is removed mid-task sequence, any subsequent tasks should be removed as well. Lastly, $C_{AS}$, $C_{\mathcal{T}S}$, and $C_{\mathcal{T}D}$ are updated corresponding to the newly available tasks, start locations, and destinations.

\subsubsection{M2M Repair}
After the Shaw Removal operator is applied to the current solution, we apply a repair operator to allocate the tasks in $\mathcal{T}_{free}$ or until every element in $C_{\mathcal{T}S}$ or $C_{\mathcal{T}D}$ is equal to 0. The process of repairing the solution is identical to the M2M algorithm detailed in the prior section. The repair operator returns a repaired task allocation, $\mathcal{A}_{new}$. 

\subsubsection{Metropolis Acceptance Criterion}

Once the M2M repair operator returns $\mathcal{A}_{new}$, we utilize a metropolis acceptance criterion \cite{kirkpatrick1983optimization, ropke2006adaptive} to either accept or reject $\mathcal{A}_{new}$ as $\mathcal{A}_{curr}$. To compare solution quality, we define a scoring function $f: \mathcal{A}\rightarrow \mathbb{R}^+$ which computes total estimated duration of the task allocation. The metropolis acceptance criterion is then defined as: 
\[
\begin{cases}
    accept & \text{if } f(\mathcal{A}_{new}) > f(\mathcal{A}_{curr}) \\
    accept & \text{else }  rand[0,1]<e^{-(f(\mathcal{A}_{new})-f(\mathcal{A}_{curr}))/T} \\
\end{cases}
\]
Here, $\mathcal{A}_{new}$ is accepted if $f(\mathcal{A}_{new}) > f(\mathcal{A}_{curr})$. Otherwise, $\mathcal{A}_{new}$ is accepted when a randomly sampled number between 0 and 1 is less than the value of the described function. Within this function, $T$ is a temperature value which is initially set to $T_0$ at the first iteration of LNS, then decreased using a decay rate $\alpha$, in which $T$ is decreased after each LNS iteration, $T=T*\alpha$. A higher value of $T$ results in a higher likelihood of accepting a worse $\mathcal{A}_{new}$ to explore.

\subsection{M2M-wSKU: Accounting for Item Distribution}
\label{sec:sku}

So far, the cost function only considers the estimated duration of $a_m$ visiting $s_p$ and $d_q$, which does not consider other important aspects of the MRTA solution, such as the distribution of items in the environment. Ideally, items should be widely distributed in the environment so as to allow agents greater options for current and future tasks and to reduce congestion in the environment. To facilitate this, we track individual items by their Stock Keeping Unit (SKU), a unique code assigned to each product, and alter M2M to incentivize assignments that remove items which are close together while encouraging destinations which place SKUs further away from existing items of the same SKU. We refer to this variant as \textit{M2M with SKU Distribution (M2M-wSKU)}. 

To incentivize choosing start locations and destinations which improve a SKU's distribution, we modify the cost function by performing a nearest neighbor search at $s_p$ or $d_q$ depending on if $\tau_n \in \mathcal{T}_{out}$ or $\tau_n \in \mathcal{T}_{in}$ respectively. 
The cost function is then modified as: 
\[
c_{m, n, p, q}=\begin{cases}
    K & \text{if } s_p \notin S_n \\
    K & \text{if } d_q \notin D_n \\
    K & \text{if } s_p \in V_{alloc} \\
    K & \text{if } d_q \in V_{alloc} \\
    w_b(cost(a_m, s_p)+cost(s_p, d_q)) \\ -w_sNearestNeighbor(s_p) & \text{if } \tau_n \in \mathcal{T}_{out} \\
    w_b(cost(a_m, s_p)+cost(s_p, d_q)) \\ +w_sNearestNeighbor(d_q) & \text{if } \tau_n \in \mathcal{T}_{in}
\end{cases}
\]
Here, $w_b$ is the base cost weight of the estimated travel duration, while $w_s$ is the SKU weight. When $\tau_n \in \mathcal{T}_{out}$, locations close to existing items of the same SKU are encouraged over ones further away. Inversely, when $\tau_n \in \mathcal{T}_{in}$, locations further from existing items of the same SKU are encouraged over ones closer. To compute the nearest neighbor search, the locations of each SKU are stored as a KD-Tree \cite{bentley1975multidimensional}, a special type of binary tree that supports nearest neighbor search in O(log(n)) time on average. In our algorithm, we maintain a KD-Tree for each type of SKU in the environment and update the tree whenever an item is placed or removed from the environment.

\subsection{Multi-Agent Path Finding}
\label{sec:mapf}

Once $\mathcal{A}_{final}$ is returned from LNS, we utilize Multi-Agent Path Finding (MAPF) algorithm to compute a corresponding motion plan $\mathcal{M}$ for the agents to execute. In this work, compute path plans segment-by-segment, rather than from $l_m(t)$ to the final destination in $\mathcal{T}^m$. We evaluated M2M and M2M-wSKU using Enhanced Conflict Based Search (ECBS) \cite{barer2014suboptimal} and Priority Based Search (PBS) \cite{ma2019searching} and found PBS performed better with our task allocation method and setting. 


\section{Experimental Setup} \label{sec:experimental setup}

In this section, we provide an overview of our experimental domain, experimental conditions, chosen model and study parameters, and metrics. 

\begin{figure}
    \centering
    \includegraphics[width=1.0\linewidth]{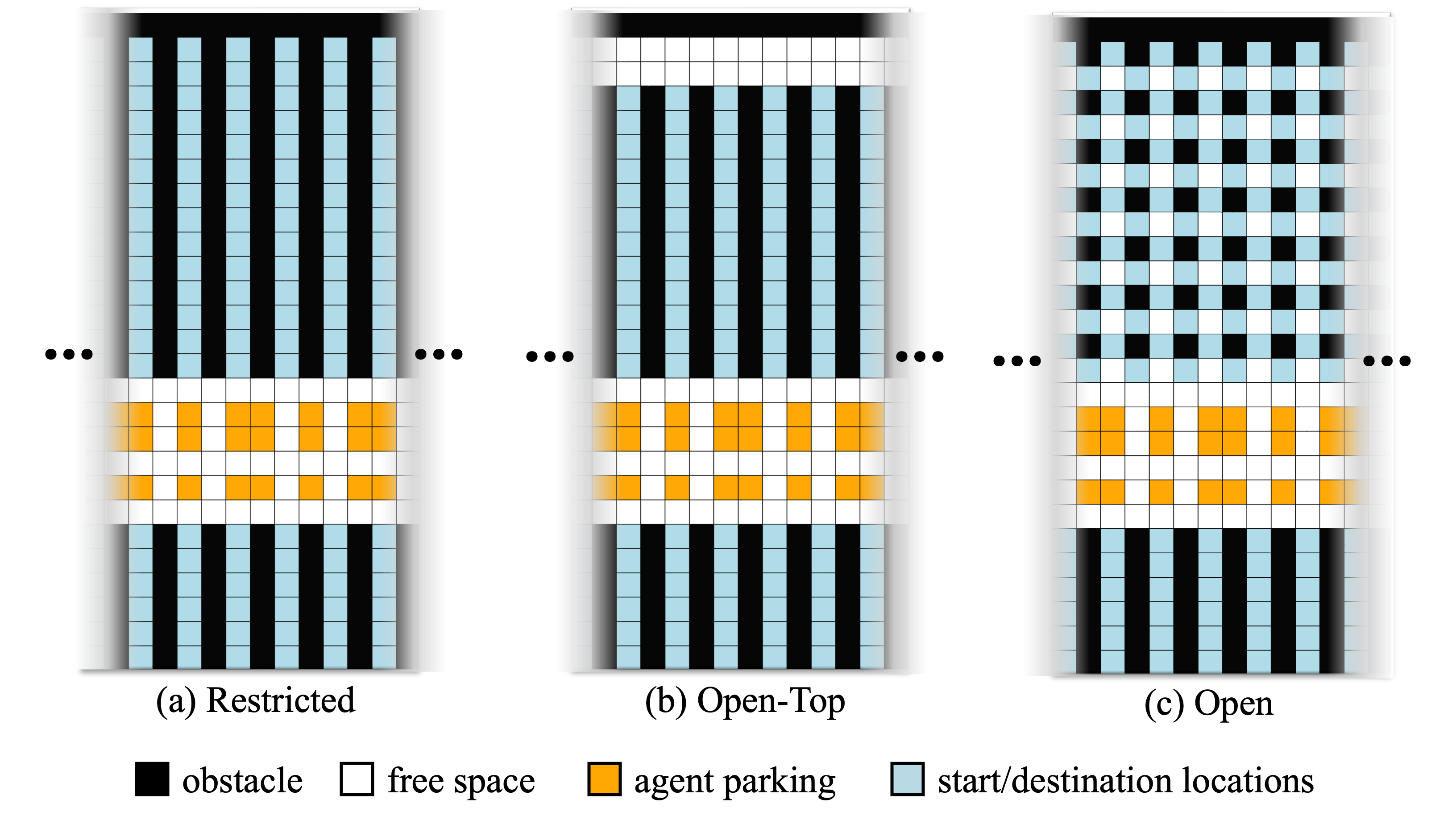}
    \caption{Partial view of our three simulated warehouse layouts. All non-obstacle regions are traversable by agents, subject to collision constraints. Each map is 27$\times$ 50.}
    \label{fig:maps}
    \vspace{-1em}
\end{figure}

\smallskip
\noindent \textit{Experimental Domain:} We evaluate our approach in a simulated warehouse environment designed to capture the complexities of real-world multi-agent pickup and delivery tasks. The warehouse is modeled as a discrete two-dimensional grid with designated storage locations, delivery stations, and narrow aisles that constrain agent movement; however, the simulation operates under a few idealized assumptions, including perfect localization of agents and SKUs, noiseless sensing, lossless communication, and no actuation errors or task failures.
We use three warehouse layouts, shown in Figure \ref{fig:maps}. The restricted map contains long aisles with a single entry point close to the item inbound/outbound loading area at the bottom. The open-top map has additional open space at the top of the map, allowing agents to enter or exit aisles at either end. The open map contains a grid layout for the inventory. All three maps mimic layouts currently in use in warehouse logistics operations.

\smallskip
\noindent \textit{Experimental Conditions:} We perform experiments comparing the following MAPD approaches.
\begin{flushitemize}
    \item \textbf{M2M:} Our Many-to-Many MAPD approach, as described in Section \ref{sec:methods}.
    \item \textbf{M2M-wSKU:} Our Many-to-Many MAPD variant incorporating SKU distribution information into the task assignment cost function, as described in Section \ref{sec:methods}. 
    \item \textbf{LNS-PBS:} As discussed in Section \ref{sec:related works}, prior work has demonstrated that LNS-PBS outperforms other prior MAPD methods \cite{lnspbs} and thus represents the state of the art in MAPD solutions. As a result, we utilize LNS-PBS as our baseline method, modifying its one-to-one pickup-and-delivery approach to operate in our many-to-many domain. Specifically, prior to running LNS-PBS, we convert each many-to-many task, (i.e. $\tau_n = (S_n, D_n)$) into a one-to-one task (i.e. $\tau_n = (s_p, d_q)$) such that we minimize the distance between $s_p$ and $d_q$ ($\arg\min_{\substack{s_p \in S_n, d_q \in D_n}} \; \text{cost}(s_p,d_q)$). Once all tasks in $\mathcal{T}_{free}$ have been converted to one-to-one tasks, we compute a task allocation and motion plan using the standard LNS-PBS approach.

\end{flushitemize}

\noindent In addition to prior work in MAPD, we investigated the use of a mixed-integer program (MIP) and set partitioning (SP) as suitable baselines. Work by Walteros et al. \cite{walteros2014integer} investigated the use of MIP and SP formulations to solve higher dimension assignment problems. For a four-dimension assignment with thirty elements in each set (a smaller scale problem than ours), MIP times out after 2 hours and SP requires 8.76 seconds to compute a solution. Neither timing is efficient enough for our large-scale warehouse use case so we do not include these techniques as baselines.

\smallskip
\noindent \textit{Parameters:} For M2M and M2M-wSKU, we set the Shaw Removal weights to $\omega_1=9$ and $\omega_2=3$ from \cite{lnspbs, ropke2006adaptive}, the initial temperature for the simulated annealing acceptance function is $T_0=1.0$ and the decay rate is $\alpha=0.99$. The total time limit for all task allocation steps (M2M and LNS combined) is set to 1 second with the number of tasks to be removed set to 3.  We set the task sequence limit to $\beta=3$. For M2M-wSKU, we performed an ablation and set the cost weights as $w_b=1.0$ and $w_s=0.25$. The LNS-PBS parameters are set the same as in Xu et al. \cite{lnspbs}.

For the simulation parameters, all maps were sized to 27$\times$50, with $M=40$ agents. Each experiment was ran for 8 wall-clock hours. At each timestep 4 tasks were released into the system with a limit of 120 total active tasks. We used 30 unique SKUs with the initial randomized inventory with a prespecified inventory density, which is maintained throughout the simulation. Performance of each experimental condition is reported as an average over 10 simulations. M2M\footnote{The code is available at https://github.gatech.edu/RAIL/M2M} and M2M-wSKU are implemented in Python, while implementation of LNS-PBS is provided by the authors in C++. All experiments were conducted on a machine equipped with an 11th Gen Intel(R) Core(TM) i7-11700K CPU @ 3.60 GHz and 32 GB DDR4 RAM; no GPU acceleration was used.

\smallskip
\noindent \textit{Metrics:} We evaluate performance using three metrics:
\begin{flushitemize}
    \item \textbf{Throughput:} calculated as the number of tasks completed per unit time. We report the mean, median, and standard deviation throughput over the simulations and experimental map conditions.
    \item \textbf{Computation Time:} is reported by the mean and std-dev computation time for each method's task allocation step. 
    \item \textbf{Scalability:} is reported by the mean and std-dev computation time for $\mathcal{A}_{initial}$ with increases values of the number of agents (M) and number of tasks (N) in the restricted map. We additionally analyize the affect of a larger map (61$\times$100) restricted map on the scalability of M2M. The reported values are averaged over 10 task allocations. 
\end{flushitemize}


\section{Results} \label{sec: results}

\begin{figure*}[t!]
    \centering
    \includegraphics[width=0.33\textwidth]{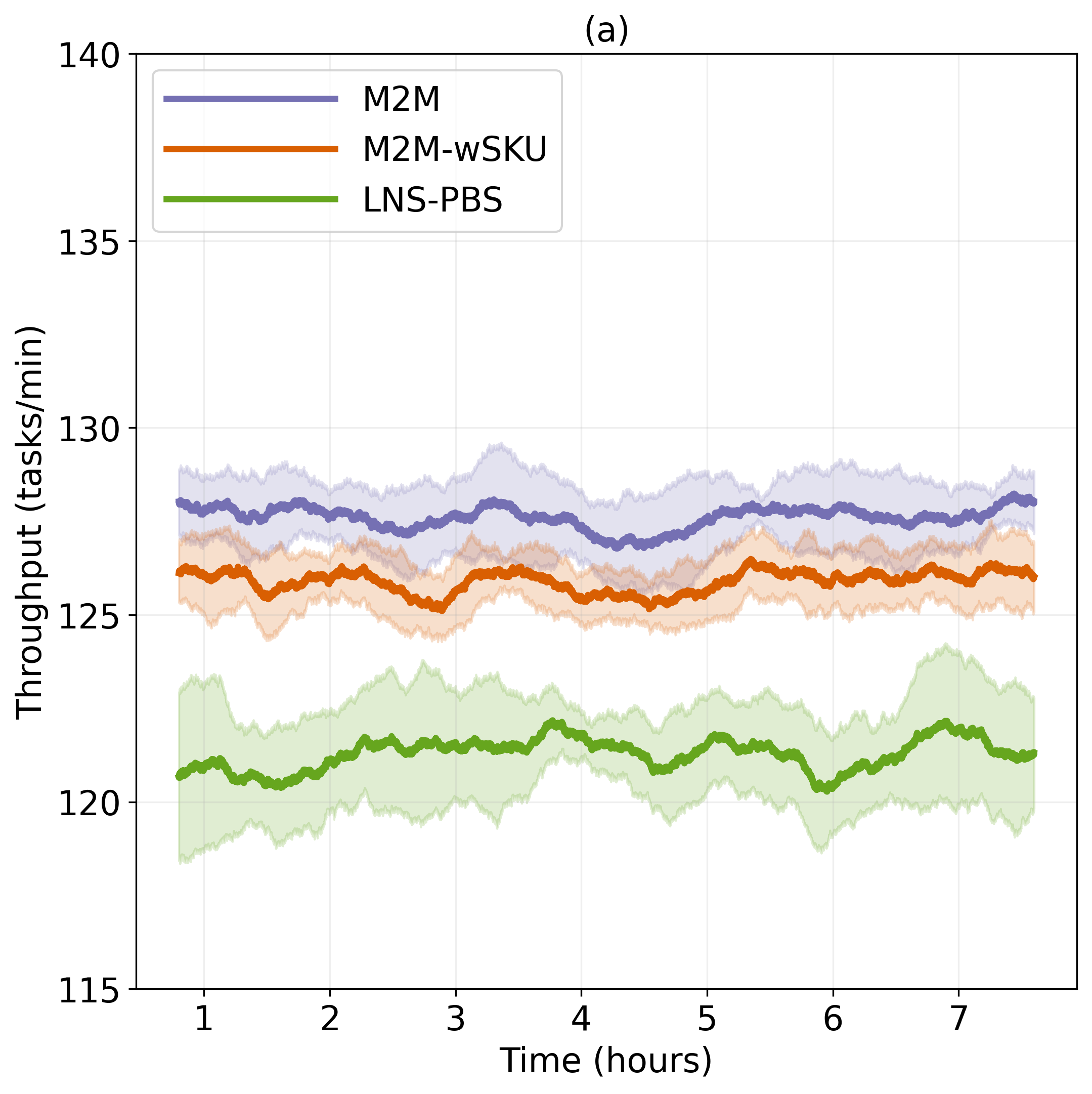}%
    \includegraphics[width=0.33\textwidth]{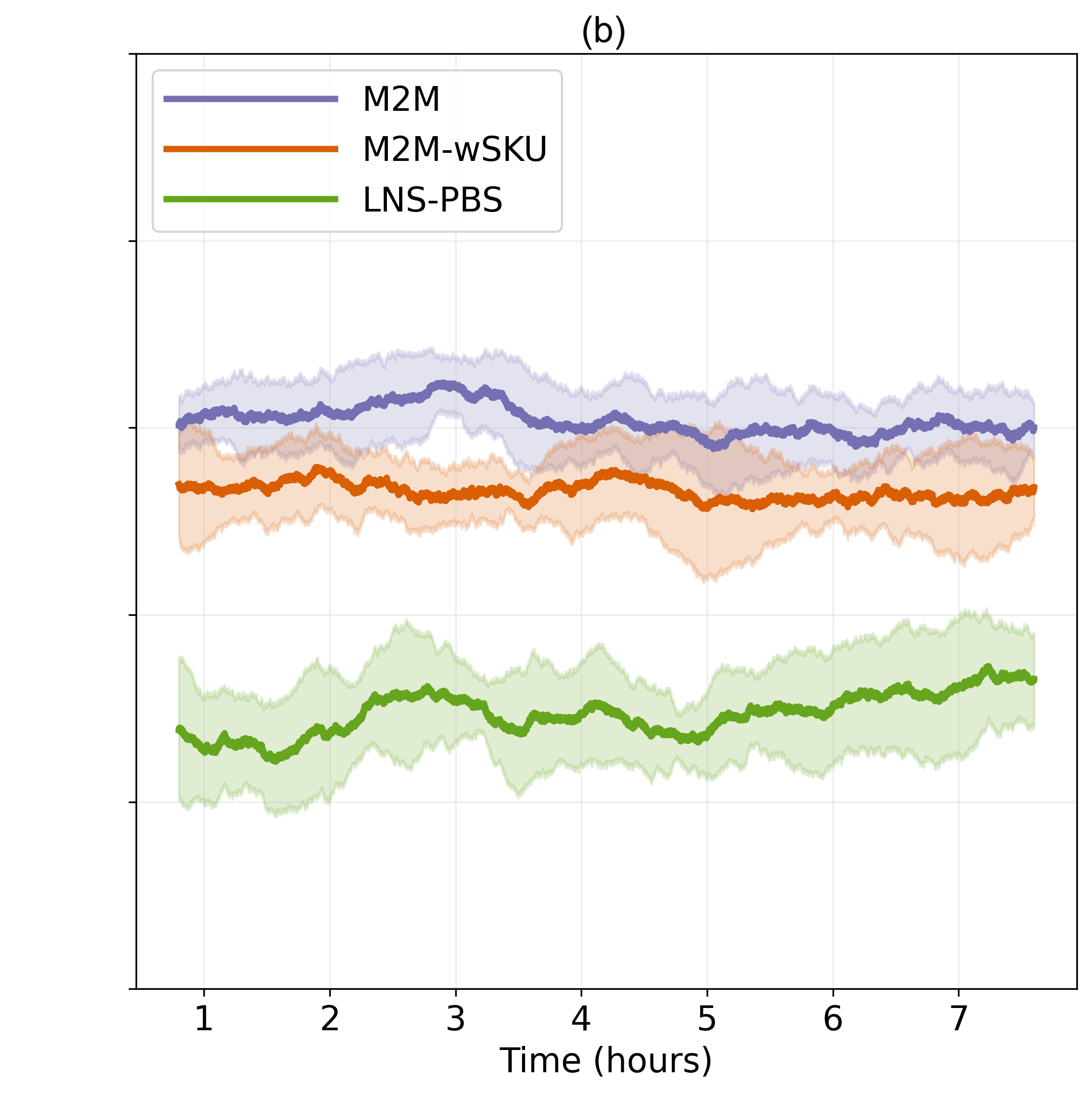}%
    \includegraphics[width=0.33\textwidth]{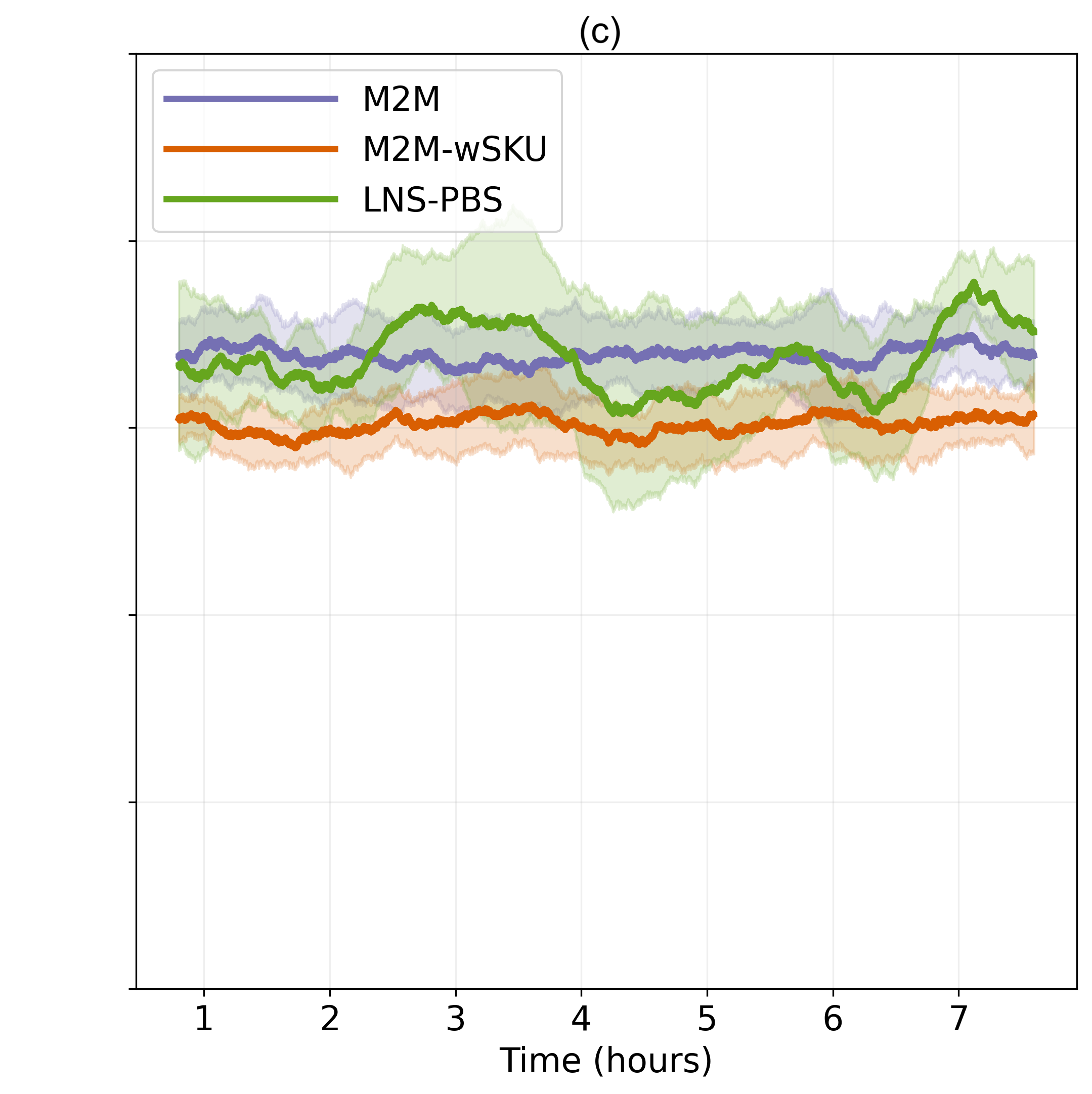}
    \caption{Comparison of the rolling mean (solid) and std-dev (shaded region) of throughput (tasks/min.) metric reported across Restricted (a), Open-Top (b), and Open (c) maps. Our method, M2M (purple), and variant, M2M-wSKU (orange), maintain a higher mean throughput over the baseline LNS-PBS (green) across all map types. Between our two variants, M2M maintains the highest mean throughput over M2M-wSKU.}
    \vspace{-1em}
    \label{fig:throughput-performance}
\end{figure*}

\begin{table*}[t!]
\centering
\begin{tabular}{|l|c|c|c|c|c|c|c|c|c|c|}
\hline
& \multicolumn{3}{c|}{Restricted Map} 
& \multicolumn{3}{c|}{Open-Top Map} 
& \multicolumn{3}{c|}{Open Map} &  \\ \hline
Method & Mean & Median & Std-Dev & Mean & Median & Std-Dev & Mean & Median & Std-Dev & Mean Compute Time (s)\\ \hline
LNS-PBS & 121.22 & 121.11 & 0.32 & 122.47 & 122.39 & 0.41 & 131.75 & 131.55 & 0.56 & 1.07\\ \hline
M2M-wSKU & 125.75 & 125.85 & 0.22 & 128.24 & 128.28 & 0.38 & 129.96 & 130.03 & \textbf{0.31} & 1.16\\ \hline

M2M & \textbf{127.53} & \textbf{127.52} & \textbf{0.16} & \textbf{130.07} & \textbf{130.08} & \textbf{0.35} & \textbf{131.78} & \textbf{131.92} & 0.34 & 1.15\\ \hline
\end{tabular}
\caption{Comparison of mean, median, and std-dev throughput across Restricted, Open-Top, and Open maps (tasks/min.) averaged over ten trials for each map and algorithm. M2M outperforms M2M-wSKU across all maps, while M2M maintains or outperforms LNS-PBS across all maps.}
\vspace{-1em}
\label{tab:throughput-performance}
\end{table*}

In this section, we report our results on the above metrics. 

\subsection{Task Throughput across Maps}

We first compare the task throughput of the baseline LNS-PBS against our M2M and M2M-wSKU variants across all three map types, in which the inventory density is set to 30\%. The results are summarized in Table \ref{tab:throughput-performance} and Figure \ref{fig:throughput-performance}. M2M consistently outperforms or maintains comparable performance against LNS-PBS across all three map types, with percent differences of 4.95\%, 6.13\%, and 0.02\% in mean throughput for the restricted, open-top, and open maps respectively. On average, this performance difference results in 3,028, 3,648, and 14 additional tasks for M2M over LNS-PBS. The results from Figure \ref{fig:throughput-performance} show that as the map layout becomes more "open", the performance of all three methods increases. In the open map setting, the performance between M2M and LNS-PBS is nearly identical, which shows that our method is able to maintain parity with prior work in more open environments while achieving higher performance in denser environments. Comparing M2M with M2M-wSKU, we observe M2M consistently outperforms M2M-wSKU across all map types, which suggests that the time for M2M-wSKU to distribute items throughout the warehouse does not result in additional performance gains. Examining the standard deviation of throughput across our experiments, we observe that all methods maintain low standard deviation across all three map types.


\textbf{Summary:} We find that M2M consistently outperforms or maintains comparable throughput against LNS-PBS across all map types, with all methods maintaining low standard deviation throughput. 

\subsection{Task Throughput across Inventory Densities}

In addition to comparing the task throughput across multiple environments, we compare the performance of M2M, M2M-wSKU, and LNS-PBS in the restricted map across 30\%, 60\%, and 90\% inventory densities, which corresponds to 3.5, 7, and 10.5 occurrences of each SKU on average at a given timestep. The results are summarized in Table \ref{tab:inventory_vary}. We find that M2M consistently outperforms LNS-PBS across all inventory densities, with an additional 4.95\%, 2.40\%, and 38.77\% increase in mean throughput for the 30\%, 60\%, and 90\% inventory densities respectively. The increase in performance for M2M corresponds to 3,028, 1,421, and 22,132 additional tasks over LNS-PBS. These results demonstrate that M2M remains resilient in highly dense warehouse inventory conditions, whereas LNS-PBS experiences significant performance degradation. We observe M2M outperforms M2M-wSKU across inventory densities, with the performance difference between the two decreasing as inventory density increases. All methods maintain low standard deviation across all levels of fullness. 

\textbf{Summary:} We find that M2M consistently outperforms M2M-wSKU and LNS-PBS across all experimental inventory densities, with all methods maintaining low standard deviation throughput. 



\begin{table*}[t!]
\centering
\begin{tabular}{|l|ccc|ccc|ccc|}
\hline
\multicolumn{1}{|c|}{} & \multicolumn{3}{c|}{Restricted (30\%)} & \multicolumn{3}{c|}{Restricted (60\%)} & \multicolumn{3}{c|}{Restricted (90\%)} \\ \hline
Method & \multicolumn{1}{c|}{Mean} & \multicolumn{1}{c|}{Median} & Std-Dev & \multicolumn{1}{c|}{Mean} & \multicolumn{1}{c|}{Median} & Std-Dev & \multicolumn{1}{c|}{Mean} & \multicolumn{1}{c|}{Median} & Std-Dev \\ \hline
LNS-PBS & \multicolumn{1}{c|}{121.22} & \multicolumn{1}{c|}{121.11} & 0.32 & \multicolumn{1}{c|}{120.63} & \multicolumn{1}{c|}{120.66} & \textbf{0.27} & \multicolumn{1}{c|}{72.82} & \multicolumn{1}{c|}{72.83} & \textbf{0.49} \\ \hline
M2M-wSKU & \multicolumn{1}{c|}{125.75} & \multicolumn{1}{c|}{125.85} & 0.22 & \multicolumn{1}{c|}{122.17} & \multicolumn{1}{c|}{122.30} & 0.91 & \multicolumn{1}{c|}{118.84} & \multicolumn{1}{c|}{118.95} & 0.59 \\ \hline
M2M & \multicolumn{1}{c|}{\textbf{127.53}} & \multicolumn{1}{c|}{\textbf{127.52}} & \textbf{0.16} & \multicolumn{1}{c|}{\textbf{123.59}} & \multicolumn{1}{c|}{\textbf{123.37}} & 0.59 & \multicolumn{1}{c|}{\textbf{118.93}} & \multicolumn{1}{c|}{\textbf{118.97}} & 0.96 \\ \hline
\end{tabular}%

\caption{Comparison of mean, median, and std-dev throughput across 30\%, 60\%, and 90\% inventory densities in the restricted map averaged over ten trials for each level of inventory density and algorithm. M2M outperforms LNS-PBS and M2M-wSKU across all inventory densities.}
\label{tab:inventory_vary}
\end{table*}



\subsection{Computation Time}
Table \ref{tab:throughput-performance} reports the mean computation time for LNS-PBS, M2M and M2M-wSKU as 1.07s$\pm$0.01s, 1.15s$\pm$0.03s, and 1.16s$\pm$0.04s, respectively. Although we limit the task allocation computation time to 1 second for each technique, additional time results from data preprocessing (e.g. converting many-to-many problem into a one-to-one for LNS-PBS) and pre- and post-process data logging.  

\textbf{Summary:} All three techniques have similarly strong computational performance.

\subsection{Scalability}
Table \ref{tab:scalability} shows the scalability of the M2M algorithm, reported as the computation time required to obtain $A_{initial}$ for domains with varying number of agents ($M$) and tasks ($N$). Our results show that M2M scales to 150 agents and 150 tasks within the 1-second compute limit window.  
Additionally, we evaluated scalability by increasing the map size from 27$\times$50 to 61$\times$100.  Under these conditions, M2M scales to only 50 agents and tasks ($\mu=1.027s;\sigma=0.149s$) compared to the 150 with the smaller restricted map. This is due to the increased number of endpoints in the environment, which increases the number of start locations and destinations for a given task. 

\textbf{Summary:} M2M (and M2M-wSKU by extension) demonstrates scalability up to 150 agents and 150 tasks in the defined map size of 27$\times$50.  Increasing map size reduces scalability to 50 agents and tasks.

\begin{table}[]
\begin{tabular}{|c|c|c|}
\hline
\begin{tabular}[c]{@{}c@{}}Number of Agents (M) \\ and Tasks (N)\end{tabular} & \begin{tabular}[c]{@{}c@{}}Mean $\mathcal{A}_{initial}$ \\ Computation Time(s)\end{tabular} & \begin{tabular}[c]{@{}c@{}}Std-Dev $\mathcal{A}_{initial}$\\  Computation Time(s)\end{tabular} \\ \hline
M=50;N=50 & 0.065s & 0.007s \\ \hline
M=70;N=70 & 0.143s & 0.012s \\ \hline
M=90;N=90 & 0.247s & 0.019s \\ \hline
M=110;N=110 & 0.390s & 0.022s \\ \hline
M=130;N=130 & 0.591s & 0.023s \\ \hline
M=150;N=150 & 0.809s & 0.043s \\ \hline
\end{tabular}
\caption{The scalability of the M2M algorithm shown with an increasing number of agents (M) and tasks (N) in the restricted map. We see that M2M scales to 150 agents and tasks given a 1 second computation budget.}
\vspace{-1em}
\label{tab:scalability}
\end{table}

\section{Conclusion} \label{sec: conclusion}

In summary, our results show that M2M consistently matches or outperforms the prior state-of-the-art method LNS-PBS across all map types. Moreover, M2M remains robust to increasing inventory density, while the mean throughput of LNS-PBS degrades significantly in high-density inventory scenarios.
Additionally, we find M2M strictly outperforms our variant M2M-wSKU, which suggests the additional time for M2M-wSKU to distribute items throughput the warehouse does not result in additional performance gains. We also demonstrate the scalability of M2M with a larger environment and a higher number of agents and tasks. In future work, we intend to improve the scalability of M2M with respect to agents, tasks, and environment sizes. 

\addtolength{\textheight}{-12cm}   



\bibliographystyle{IEEEtranS}
\bibliography{references}

\end{document}